\documentclass[letterpaper]{article}
\usepackage{tech}
\usepackage{amsmath}
\usepackage{graphicx}
\usepackage{url}
\usepackage{enumitem}
\def\one{\mathbf{1}}
\setlist[enumerate]{itemsep=0mm}
\begin{document}
%


\title{Box Drawings for Learning with Imbalanced Data}
%
%
%
%
%

\author{Siong Thye Goh \and Cynthia Rudin \\Massachusetts Institute of Technology\\
    Cambridge, MA 02139, USA \\}
\maketitle

\date{12 February 2014}

\maketitle
\begin{abstract}
The vast majority of real world classification problems are imbalanced, meaning there are far fewer data from the class of interest (the positive class) than from other classes. We propose two machine learning algorithms to handle highly imbalanced classification problems. The classifiers are disjunctions of conjunctions, and are created as unions of parallel axis rectangles around the positive examples, and thus have the benefit of being interpretable. The first algorithm uses mixed integer programming to optimize a weighted balance between positive and negative class accuracies. Regularization is introduced to improve generalization performance. The second method uses an approximation in order to assist with scalability. Specifically, it follows a \textit{characterize then discriminate} approach, where the positive class is characterized first by boxes, and then each box boundary becomes a separate discriminative classifier. This method has the computational advantages that it can be easily parallelized, and considers only the relevant regions of feature space. 
\end{abstract}



\section{Introduction}
Our interest is in deriving interpretable predictive classification models for use with imbalanced data. 
Data classification problems having imbalanced (also called ``unbalanced") class distributions appear in many domains, ranging from mechanical failure detection or fault detection, to fraud detection, to text and image classification, to medical disease prediction or diagnosis. Imbalanced data cause typical machine learning methods to produce trivial results, that is, classifiers that only predict the majority class. One cannot optimize vanilla classification accuracy and use standard classification methods when working with imbalanced data. This is explained nicely by Chawla, Japkowicz, and Kolcz  \cite{ChawlaEtAl}  
who write: ``\textit{The class imbalance problem is pervasive and ubiquitous, causing trouble to a large segment of the data mining community.}"

\begin{figure}
\vspace{0.5cm}
	\centering
		\includegraphics[width=8cm]{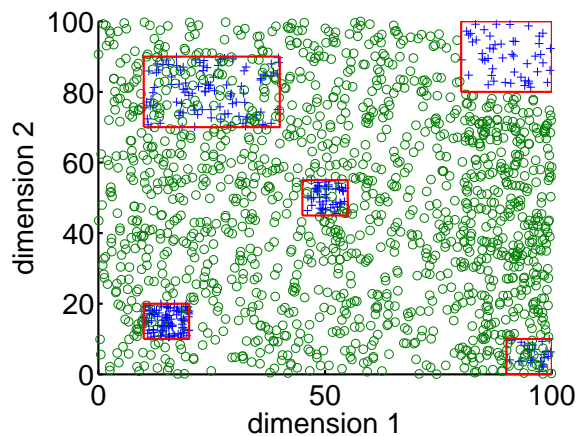}
	\caption{Example of box drawing classifier.}
	\label{fig:exampleofdata}
\end{figure}

In order for the models we derive to be interpretable to human experts, our classifiers are formed as a union of axis parallel rectangles around the positive (minority class) examples, and we call such classifiers \textit{box drawing classifiers}. These are ``disjunctions of conjunctions" where each conjunction is a box. An example of a box drawing classifier we created is in Figure \ref{fig:exampleofdata}, exemplifying our goal to classify the positive examples correctly even if they are scattered within a sea of negative examples. Our classifiers are regularized in several ways, to prefer fewer boxes and larger boxes. We take two polar approaches to creating box drawing classifiers, where the first is an exact method, based on mixed integer programming (MIP). This method, called \textit{Exact Boxes} can be used for small to medium sized datasets, and provides a gold standard to compare with. If we are able to make substantial approximations and still obtain performance close to that of the gold standard, our approximations would be justified. Our second method, \textit{Fast Boxes} makes such an approximation. 

Fast boxes takes the approach of \textit{characterize then discriminate}, where we first characterize the positive (minority) class alone, and then bring in the negative examples to form decision boundaries around each clusters of positives. This approach has significant computational advantages, in that using just the minority class in the first step requires a small fraction of the data, assuming a high imbalance ratio. Also by creating decision boundaries locally in the second step, the number of examples involved in each classifier is smaller; further, creating each classifier separately allows computations to be made parallel, though since the computation for each decision boundary is analytical, that may not be necessary for many datasets. The computation is analytical because there is a closed form solution for the placement of the decision boundary. Thus, the discriminate step becomes many parallel local analytical calculations. This is much simpler and scalable than, for instance, a decision tree that chooses splits greedily and fails to scale with dimension and large number of observations.

We make several experimental observations, namely that: box drawing classifiers become more useful as data imbalance increases; the approximate method performs at the top level of its competitors, despite the fact that it is restricted to producing interpretable results; and performance can be improved on the same datasets by using the mixed integer programming method. 

After related work just below, we describe the advantages of our approach in Section 2. In Section 3, we introduce our two algorithms. Experimental results will be presented in Section 4. Section 4 provides a vignette to show how box drawing models can be interpretable. In Section 5, theoretical generalization bounds will be presented for box drawing classifiers. Section 6 discusses possible approaches to make the MIP formulation more scalable.

\section{Related Works}

Overviews of work on handling class imbalance problems include those of He and Garcia \cite{HHEG}, Chawla, Japkowiz and Kolcz \cite{ChawlaEtAl} and Qi \cite{Y04}. Many works discuss problems caused by class imbalance \cite{Weissalone,Prati}. There are many avenues of research that are not directly related to the goal of interpretable imbalanced classification, specifically kernel and active learning methods \cite{Raskutti,Wu}, and work on sampling \cite{Abe,SMOTE} that includes undersampling, oversampling, and data generation, which can be used in conjunction with methods like the ones introduced here. We use a cost-sensitive learning approach in our methods, similar to Liu and Zhou \cite{LiuZhou2} and McCarthy et al$.$ \cite{McCarthy}. We note that many papers on imbalanced data do not experimentally compare their work with the cost-sensitive versions of decision tree methods. We choose to compare with other cost-sensitive versions of decision trees as our method is a cost-sensitive method. 

There is some evidence that more complex approaches that layer different learning methods seem to be helpful for learning \cite{Raskutti,Wu}, though the results would not be interpretable in that case. This, however, is in contrast with other views (e.g., \cite{simplerule}) that for most common datasets, simple rules exists and we should explore them. 

The works most similar to ours are that of the Patient Rule Induction Method (PRIM) \cite{Friedman} and decision tree methods for imbalanced classification (e.g., \cite{Japkowicz}), as they partition the input space like our work. Approaches that partition space tend to recover simple decision rules that are easier for people to understand. Decision tree methods are composed using greedy splitting criteria, unlike our methods. PRIM is also a greedy method that iteratively peels off parts of the input space, though unfortunately we found it to be extremely slow - as described by Sniadecki \cite{Sniadecki}, ``PRIM is eloquently coined as a patient method due to the slow, stepwise mechanism by which it processes the data." Neither our Exact Boxes nor Fast Boxes methods are greedy methods, though Fast Boxes makes a different type of approximation, which is to characterize before discriminating. As discussed by Raskutti \cite{Raskutti}, one-class learning can be useful for highly imbalanced datasets - our characterization step is a one-class learning approach.


\section{New Algorithms}

We start with the mixed-integer programming formulation, which acts as our gold standard for creating box drawing classifiers when solved to optimality. 

\subsection{Exact Boxes}
For box drawing classifiers, a minority class (positive) example is correctly classified only if it resides within at least one box. A majority class (negative) example is correctly classified if it does not reside in any box. We are given training examples $\{(\mathbf{x}_i,y_i)\}_{i=1}^m, \mathbf{x}_i\in \mathcal{R}^n, y_i\in\{-1,+1\}$.
We introduce some notation in Table \ref{MIPnotation} that we will use throughout this subsection. We use this notation from here on.

\begin{table}[ht]
\centering
\small
\begin{tabular}{|c|c|}
\hline 
Notation & Definitions \\ \hline
$K$ & Number of parallel axes boxes \\ \hline
$m$ & Number of examples \\ \hline
$n$ & Number of features \\ \hline
$i$ & Index for examples \\ \hline
$j$ & Index for features \\ \hline
$x_{ij}$ & $j$-th feature of example $i$ \\ \hline
$k$ & Index for box \\ \hline
$l_{jk}$ & Lower boundary of feature $j$ for box $k$ \\ \hline
$u_{jk}$ & Upper boundary of feature $j$ for box $k$ \\ \hline
$v$ & Margin for decision boundary  \\ \hline
$\widetilde{l}_{ijk}$ & $\widetilde{l}_{ijk}=1$ if $x_{ij} > l_{jk}+v$ and $0$ otherwise \\ \hline
$\widetilde{u}_{ijk}$ & $\widetilde{u}_{ijk}=1$ if $x_{ij} < u_{jk}-v$ and $0$ otherwise \\ \hline
$w_{ik}$ & $w_{ik}=1$ if example $i$ is in box $k$ and $0$ otherwise \\ \hline
$z_i$ & $z_i=1$ if it is classified correctly. \\ \hline
$S_+$ & Index set of example of minority class \\ \hline
$S_-$ & Index set of example of majority class \\ \hline
$c_e$ & A regularizer to encourage expansion of box \\ \hline
$c_I$ & Weight for majority class, $c_I<1$ \\ \hline
\end{tabular}
\caption{Notation for Box Drawings with Mixed Integer Programming}
\label{MIPnotation}
\end{table} 

The \textit{Exact Boxes} method solves the following, where the hypothesis space $\mathcal{F}$ is the set of box drawings (unions of axis parallel rectangles), where $f\in\mathcal{F}$ has $f:\mathcal{R}^n\rightarrow \{-1,1\}$.
\begin{eqnarray*}
\max_{f \in \mathcal{F}} \sum_{i:y_i=1} \one_{[f(\mathbf{x}_i)=1]} + C_I \sum_{i:y_i=-1} \one_{[f(\mathbf{x}_i)=-1]} \\ - C_E(\# \textrm{of boxes of } f).
\end{eqnarray*}
The objective is a weighted accuracy of positives and negatives, regularized by the number of boxes.
This way, the number of boxes is not fixed, and a smaller number of clusters is preferred (analogous to nonparametric Bayesian models where the number of clusters is not fixed). Our gold standard will be the minimizer of this objective. We now derive the MIP that computes this minimizer.

If $i \in S_+$, the definitions of $\widetilde{l}_{ijk}$, $\widetilde{u}_{ijk}$, $w_{ik}$, and $z_i$ give rise to the following constraints:
\begin{eqnarray}
l_{jk}+v<x_{ij} ~\mbox{iff}~ \widetilde{l}_{ijk}=1\label{my1}\\
u_{jk}-v>x_{ij} ~\mbox{iff}~ \widetilde{u}_{ijk}=1,\label{my2}
\end{eqnarray}
which say that $x_{ij}$ need to be at least margin $v$ away from the lower (resp. upper) boundary of the box in order for  $\widetilde{l}_{ijk}=1$ (resp.  $\widetilde{u}_{ijk}=1$). Further, our definitions give rise also to
\begin{equation}\label{my3}
\sum_{j=1}^n{\widetilde{u}_{ijk}+\widetilde{l}_{ijk}}>2n-1~\mbox{iff}~ w_{ik}=1,
\end{equation}
which says that for example $i$ to be in box $k$, all of the $\widetilde{u}_{ijk}$ and $\widetilde{l}_{ijk}$ are 1 for box $k$. We also have, still for $i\in S_+$, that the example must be in one of the boxes in order to be classified correctly, that is: 
\begin{equation}\label{my4}
\sum_{k=1}^Kw_{ik}>0 ~\mbox{iff}~ z_i=1.
\end{equation}
Continuing this same type of reasoning for $i \in S_-$, the definitions of $\widetilde{l}_{ijk}$ ,$\widetilde{u}_{ijk}$, $w_{ik}$, and $z_i$ give rise to the following constraints:
\begin{eqnarray}
l_{jk}-v>x_{ij}&\mbox{iff}&\widetilde{l}_{ijk}=1\nonumber\\
u_{jk}+v<x_{ij}&\mbox{iff}&\widetilde{u}_{ijk}=1\nonumber\\
\sum_{j=1}^n{\widetilde{u}_{ijk}+\widetilde{l}_{ijk}}>0&\mbox{iff}&w_{ik}=0\nonumber\\
\sum_{k=1}^Kw_{ik}>0 &\mbox{iff}& z_i=0.\nonumber
\end{eqnarray}
By setting $M$ to be a large positive constant and setting $\epsilon$ to be a small positive number (to act as a strict inequality), we now have the following formulation:
$$\max_{{l,\widetilde{l},u,\widetilde{u},w,z}}  \left[-c_eK+ \sum_{i \in S_+} z_i+c_I \sum_{i \in S_-} z_i\right] \mbox{subject to}$$ 

\vspace{-0.5cm}
\begin{align}
\label{my5}
  x_{ij}-l_{jk}-v &\leq M \widetilde{l}_{ijk}, \forall i \in S_+, \forall j,k \\
\label{my6}
 M(\widetilde{l}_{ijk}-1)+\epsilon &\leq x_{ij}-l_{jk}-v, \forall i \in S_+, \forall j,k \\
\label{my7}
 u_{jk}-v-x_{ij} &\leq M \widetilde{u}_{ijk}, \forall i \in S_+, \forall j,k \\
\label{my8}
 M(\widetilde{u}_{ijk}-1)+\epsilon &\leq u_{jk}-x_{ij}-v, \forall i \in S_+, \forall j,k 
\end{align}
\vspace{-0.8cm}
\begin{eqnarray}
\label{my9}
\sum_{j=1}^n \widetilde{l}_{ijk}+\sum_{j=1}^n \widetilde{u}_{ijk}-2n+1 \leq w_{ik}, \forall i \in S_+, \forall j,k 
\end{eqnarray}
\vspace{-0.6cm}
\begin{eqnarray}
\label{my10}
2nw_{ik}\leq \sum_{j=1}^n \widetilde{l}_{ijk}+\sum_{j=1}^n \widetilde{u}_{ijk}, \forall i \in S_+, \forall j,k
\end{eqnarray}
\vspace{-0.6cm}
\begin{align}
\label{my11}
 \sum_{k=1}^K w_{ik}&\leq Kz_i, \forall i \in S_+, \forall k\\[-0.5cm]
\label{my12}
 z_i &\leq \sum_{k=1}^K w_{ik}, \forall i \in S_+, \forall k\\
\label{my13}
 l_{jk}-v-x_{ij} &\leq M \widetilde{l}_{ijk}, \forall i \in S_-, \forall j,k\\
\label{my14}
 M(\widetilde{l}_{ijk}-1)+\epsilon &\leq l_{jk}-v-x_{ij}, \forall i \in S_-, \forall j,k\\
\label{my15}
 x_{ij}-u_{jk}-v &\leq M \widetilde{u}_{ijk}, \forall i \in S_-, \forall j,k \\
\label{my16}
 M(\widetilde{u}_{ijk}-1)+\epsilon &\leq x_{ij}-u_{jk}-v, \forall i \in S_-, \forall j,k
\end{align}
\vspace{-0.8cm}
\begin{eqnarray}
\label{my17}
\sum_{j=1}^n \widetilde{l}_{ijk}+\sum_{j=1}^n \widetilde{u}_{ijk}-2n+1 \leq 2n(1-w_{ik}), \nonumber\\[-0.4cm]
\forall i \in S_-, \forall j,k \\
\label{my18}
 1-w_{ik} \leq \sum_{j=1}^n \widetilde{l}_{ijk}+\sum_{j=1}^n \widetilde{u}_{ijk}, \forall i \in S_-, \forall j,k
\end{eqnarray}
\vspace{-0.9cm}
\begin{eqnarray}
\label{my19}
 \sum_{k=1}^K w_{ik} &\leq& K(1-z_i),\forall i \in S_-, \forall k \\[-0.4cm]
\label{my20}
 1-z_i &\leq& \sum_{k=1}^K w_{ik},\forall i \in S_-, \forall k
\end{eqnarray}
\vspace{-0.5cm}
\begin{eqnarray}
\label{my21}
 l_{jk} \leq u_{jk}, \forall j,k.
\end{eqnarray}

Here, (\ref{my5}) and (\ref{my6}) are derived from (\ref{my1}), (\ref{my7}) and (\ref{my8}) are derived from (\ref{my2}), (\ref{my9}) and (\ref{my10}) are derived from (\ref{my3}), (\ref{my11}) and (\ref{my12}) are derived from (\ref{my4}), equations (\ref{my13})-(\ref{my20}) are derived analogously for $S_-$. The last constraint (\ref{my21}) is to make sure that the solution that we obtain is not degenerate, where the lower boundary is above the upper boundary. 
In practice, $M$ should be chosen as a fixed large number and $\epsilon$ should be chosen as a fixed small number based on the representation of numbers in the computing environment.

In total, there are $O(mnK)$ equations and $O(mnK)$ variables, though the full matrix of variables corresponding to the mixed integer programming formulation is sparse since most boxes operate only on a small subset of the data. This formulation can be solved efficiently for small to medium sized datasets using MIP software, producing a gold standard box drawing classifier for any specific number of boxes (determined by the regularization constant). The fact that Exact Boxes produces the best possible function in the class of box drawing classifiers permits us to evaluate the quality of Fast Boxes, which operates in an approximate way on a much larger scale.


\subsection{Fast Boxes}
Fast Boxes uses the approach of \textit{characterize then discriminate}. In particular, we hypothesize that the data distribution is such that the positive examples cluster together relative to the negative examples. This implies that a reasonable classifier might first cluster the positive examples and then afterwards discriminate positives from negatives. The discrimination is done by drawing a high dimensional axis-parallel box around each cluster and then adjusting each boundary locally for maximum discriminatory power. If the cluster assumption about the class distributions is not correct, then Fast Boxes could have problems, though it does not seem to for most real imbalanced datasets we have found, as we show in the experiments. Fast Boxes has three main stages as follows.

\hrulefill
\begin{enumerate}
\item Clustering stage: Cluster the minority class data into $K$ clusters, where $K$ is an adjustable parameter. The decision boundaries are initially set as tight boxes around each of the clusters of positive examples.
\item Dividing space stage: The input space of the data is partitioned to determine which positive and negative examples will influence the placement of each decision boundary. 
\item Boundary expansion stage: Each boundary is expanded by minimizing an exponential loss function. The solution for the decision boundary is analytical. 
\end{enumerate}
\hrulefill

Details of each stage are provided below.

\subsubsection{Clustering Stage}

In the clustering stage, the minority class data are clustered into $K$ clusters. Since this step involves only the minority class data, it can be performed efficiently, particularly if the data are highly imbalanced. Cross-validation or other techniques can be used to determine $K$. In our experiments, we used the basic $k$-means algorithm with Euclidean distance. Other clustering techniques or other distance metrics can be used.

After the minority class data are separated into small clusters, we construct the smallest enclosing parallel axes rectangle for each cluster. The smallest enclosing parallel axes rectangle can be computed by taking the minimum and maximum of the minority class data in each cluster and for each feature. Let $l_{s,j,k}$ and $u_{s,j,k}$ denote the lower boundary and upper boundary for the $j$-th dimension, for the $k$-th cluster. Here the subscript $s$ is for ``starting" boundary, and in the next part we will created a ``revised" boundary which will be given subscript $r$. The ``final" boundary will be given subscript $f$.

\subsubsection{Dividing Space Stage}

Define the set $X_{l,j,k}$ as follows:
\begin{eqnarray}\nonumber
X_{l,j,k}&:=&\left\{ x: x_j\leq l_{s,j,k}\right\} \cup \left\{ x: l_{s,j,k}\leq x_j  \right.
\\
&& \leq \frac{l_{s,j,k}+u_{s,j,k}}{2},\left. l_{s,p,k}\leq x_p \leq u_{s,p,k}, \right. \nonumber
\\
&& \left. p \neq j\right\}, \label{lowerpartition}
\end{eqnarray}
These are the data points that will be used to adjust the lower boundary of the $j$-th dimension of the $k$-th rectangle.

Similarly, we let 
\begin{eqnarray}\nonumber
X_{u,j,k}&:=&\left\{ x: x_j\geq u_{s,j,k}\right\} \cup \left\{ x: \frac{l_{s,j,k}+u_{s,j,k}}{2} \leq x_j    \right.\\
&& \leq u_{s,j,k}, \left. l_{s,p,k}\leq x_p \leq u_{s,p,k}, p \neq j \vphantom{\frac{l_{s,j,k}+u_{s,j,k}}{2} }\right\}.
\label{upperpartition}
\end{eqnarray}
These are the training examples that will be used to determine the upper boundary of the $j$-th dimension of the $k$-th rectangle.

\begin{figure}[h]
\vspace{0.5cm}
	\centering
		\includegraphics[width=6cm]{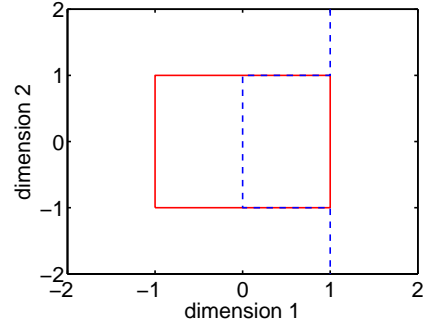}
	\caption{The examples used to determine the right vertical decision boundary are on the right side of the blue dotted line.}
	\label{fig:spacedivide}
\end{figure}

Figure \ref{fig:spacedivide} illustrates the domain for $X_{u,j,k}$ to the right of the blue dashed line.

Note that this method is very parallelizable after the clustering stage. The dividing space stage computations can be done in parallel for each cluster, and for the boundary expansion stage discussed below, each boundary of each box can be determined in parallel. 

\subsubsection{Boundary Expansion Stage}

In this stage we discriminate between positives and negatives by creating a 1-dimensional classifier for each boundary of each box. We use a regularized exponential loss. Specifically, for lower boundary $j$ of box $k$,  We minimize the following with respect to $l_{r,j,k}$ where $l_{r,j,k}$ refers to the lower boundary of the $j$-th dimension of $k$-th revised box being determined by the loss function
: 
\begin{eqnarray*}
 \lefteqn{\sum_{x \in S_+^k \cap X_{l,j,k}} \exp [{-(x_j-l_{r,j,k})}]}\\
&+&c\sum_{x \in S_-^k \cap X_{l,j,k}} \exp\left[+\left(\vphantom{\sum_{p \neq j} (\left\lfloor x_p-u_{s,p,k} \right\rfloor_+ + \left\lfloor l_{s,p,k}-x_p \right\rfloor_+)}x_j-l_{r,j,k} \right. \right.\\
&+& \left.\left.\sum_{p \neq j} (\left\lfloor x_p-u_{s,p,k} \right\rfloor_+ + \left\lfloor l_{s,p,k}-x_p \right\rfloor_+)\right)\right]+\beta l_{r,j,k}.
\end{eqnarray*}
\noindent where $c$ is the weight for the majority class, $c<1$, $S_+^k$ is the set of positive examples in the $k$-th cluster, $S_-^k$ is the set of examples not in the $k$-th cluster, $\beta$ is a regularization parameter that tends to expand the box, and $\left\lfloor . \right\rfloor$ denotes $\max(.,0)$. 
For simplicity, we use the same parameter to control the expansion for all the clusters and all the features. 
Note that the term $\sum_{p \neq j} (\left\lfloor x_p-u_{s,p,k} \right\rfloor_+ + \left\lfloor l_{s,p,k}-x_p \right\rfloor_+)$ is designed to give less weight to the points that are not directly opposite the box edges (the points that are diagonally away from the corners of the box). To explain these terms, recall that the exponential loss in classification usually operates on the term $y_if(x_i)$, where the value of $f(x_i)$ can be thought of as a distance to the decision boundary. In our case, for the lower decision boundary we use the perpendicular distance to the decision boundary $|x_j-l_{r,j,k}|$, and include the additional distance in every other dimension $p$ for the diagonal points. For the upper diagonal points we include the distance to the upper boundary $u_{s,p,k}$, namely $x_p-u_{s,p,k}$, and analogously for the points on the lower diagonal we include distance $l_{s,p,k}-x_p$. We perform an analogous calculation for the upper boundary.

Note that we perform a standard normalization of all features to be between -1 and 1 before any computation begins, which also mitigates numerical issues when dealing with the (steep) exponential loss. Another mechanism we use for avoiding numerical problems is by multiplying each term in the objective by $\exp(1)$ and dividing each term by the same factor. We will construct the derivation of the lower boundary as follows. We rewrite the objective to minimize:
\begin{align}\nonumber \label{lowerclearequation}
&R_+^{l,j,k}\exp{(-l_{s,j,k}+1+l_{r,j,k})}\\&+cR_-^{l,j,k}\exp{(l_{s,j,k}-1-l_{r,j,k})}+\beta l_{r,j,k},
\end{align}
where
\begin{eqnarray}
\label{lowersum1}
R_+^{l,j,k}&:=&\sum_{x \in S_+^k \cap X_{l,j,k}} \exp\left[{-(x_j-l_{s,j,k}+1)}\right], \\
R_-^{l,j,k}&:=&\sum_{x \in S_-^k \cap X_{l,j,k}} \exp\left[\vphantom{\sum_{p \neq j} (\left\lfloor x_p-u_{s,p,k} \right\rfloor_+ + \left\lfloor l_{s,p,k}-x_p \right\rfloor_+)}x_j-l_{s,j,k}+1\right. \nonumber \\
&&+\left.\sum_{p \neq j} (\left\lfloor x_p-u_{s,p,k} \right\rfloor_+ \right. \nonumber\\
&&+ \left.\left\lfloor l_{s,p,k}-x_p \right\rfloor_+)\vphantom{\sum_{p \neq j} (\left\lfloor x_p-u_{s,p,k} \right\rfloor_+ + \left\lfloor l_{s,p,k}-x_p \right\rfloor_+)}\right].
\label{lowersum}
\end{eqnarray} 
Because of the factors of 1 added and subtracted in the exponent, we ensure $R_+^{l,j,k}$ is at least $\exp(-1)>0.3$, avoiding numerical problems. From there, we can solve for $l_{r,j,k}$ by taking the derivative of the objective and equating it to zero. Then we multiply both sides of the resulting equation by $\exp{(l_{s,j,k}-1-l_{r,j,k})}$ and solve a quadratic equation. The the result is below. The details have been omitted due to space constraints.

\newtheorem{proposition}{Proposition}
\begin{proposition} \label{proposition1}
If $R_-^{l,j,k}>0$, the solution to (\ref{lowerclearequation}) is \begin{equation}
l_{r,j,k}=l_{s,j,k}-1+\log \left(\frac{-\beta+\sqrt{\beta^2+4cR_+^{l,j,k}R_-^{l,j,k}}}{2R_+^{l,j,k}}\right).
\label{lowerloss} 
\end{equation}. 
\end{proposition}

If $R_-^{l,j,k}=0$ or close to zero, which can happen when there are no points outside the smallest enclosing box in direction $j$, we set  $l_{r,j,k}=\overline{l_j}$ where $\overline{l_j}$ is the smallest value of feature $j$. In that case, the boundary effectively disappears from the description of the classifier, making it more interpretable. 

The interpretation of the proposition is that the boundary has moved from its starting position $l_{s,j,k}$ by amount $1-\log \left(\frac{-\beta+\sqrt{\beta^2+4cR_+^{l,j,k}R_-^{l,j,k}}}{2R_+^{l,j,k}}\right)$. 

Similarly, we let $u_{r,j,k}$ be the revised upper boundary of the $j$th dimension for the $k$-th revised box and it can be computed as follows. 

\begin{proposition}
If $R_-^{l,j,k}>0$, 
\begin{equation}
u_{r,j,k}=u_{s,j,k}+1+\log \left(\frac{\beta+\sqrt{\beta^2+4cR_+^{u,j,k}R_-^{u,j,k}}}{2cR_-^{u,j,k}}\right)
\label{upperloss}
\end{equation}
\end{proposition}
where
\begin{eqnarray}
\label{uppersum1}
R_+^{u,j,k}&:=&\sum_{x \in S_+^k \cap X_{u,j,k}} \exp\left[{-(u_{s,j,k}-x_j+1)}\right], \\
R_-^{u,j,k}&:=&\sum_{x \in S_-^k \cap X_{u,j,k}} \exp\left[ \vphantom{\sum_{p \neq j} (\left\lfloor x_p-u_{s,p,k} \right\rfloor_+}u_{s,j,k}-x_j+1 \right. \nonumber\\
&&+\left.\sum_{p \neq j} (\left\lfloor x_p-u_{s,p,k} \right\rfloor_+ \right. \nonumber \\&&+ \left. \left\lfloor l_{s,p,k}-x_p \right\rfloor_+) \vphantom{\sum_{p \neq j} (\left\lfloor x_p-u_{s,p,k} \right\rfloor_+}\right].
\label{uppersum}
\end{eqnarray}

%

The proof and interpretation are similar to Proposition \ref{proposition1}.

If $R_-^{u,j,k}=0$ or close to zero, we set $v=\overline{u_j}$ where $\overline{u_j}$ is the largest possible value for feature $j$. 

After we learn each of the decision boundaries, we perform a final adjustment that accomplishes two tasks: (i) it ensures that the box always expands rather than contracts, (ii) further expands the box to $\epsilon$ away from the nearest negative example. This gives us final values $l_{f,j,k}$ and $u_{f,j,k}$, where subscript ``$f$" is for final. Written out, this is:
\begin{eqnarray}\nonumber\label{expansion1}
l_{f,j,k}&:=&\sup\left\{ x_j| x \in S_-, x_j \right.\\&<& \left.\min(l_{r,j,k},l_{s,j,k})\right\}+\epsilon, \forall j,k \\
\nonumber\label{expansion}
u_{f,j,k}&:=&\inf\left\{ x_j| x \in S_-, x_j \right.\\&>& \left.\max(u_{r,j,k},u_{s,j,k})\right\}-\epsilon, \forall j,k
\end{eqnarray}
where $\epsilon$ is a small number. 
The boxes always expand for this algorithm, which implies that this algorithm is meant for applications where correct classification of the minority class data is crucial in practice.
This expansion step can be omitted if desired, for instance if misclassifying the negative examples is too costly.

The algorithm is summarized as follows:
\subsubsection{Overall Algorithm}

\noindent Input: number of boxes $K$, tradeoffs $c$ and $\beta$, Data $\{\mathbf{x_i},y_i\}_i$.\\
Output: Boundaries of boxes.

\begin{enumerate}
\item Normalize the features to be between -1 and 1.
\item Cluster the minority class data into $K$ clusters.
\item Construct the minimal enclosing box for each cluster, that is compute starting boundaries $l_{s,j,k}$ and $u_{s,j,k}$, the $j$-th dimension lower boundary and upper boundary respectively for the $k$'th cluster.
\item Construct data for local classifiers $X_{l,j,k}$ and $X_{u,j,k}$ based on equations (\ref{lowerpartition}) and (\ref{upperpartition}) respectively.
\item Compute $R_+^{l,j,k}$, $R_-^{l,j,k}$, $R_+^{u,j,k}$, $R_-^{u,j,k}$, according to equations (\ref{lowersum1}), (\ref{lowersum}), (\ref{uppersum1}), and (\ref{uppersum}).
\item Compute $l_{r,j,k}$ based on equation (\ref{lowerloss}) and $u_{r,j,k}$ based on equation (\ref{upperloss}) respectively.
\item Perform expansion based on equations (\ref{expansion1}) and (\ref{expansion}). 
\item Un-normalize by rescaling the features back to get meaningful values.
\end{enumerate}
Note that after the clustering step on the minority class data, all the other steps are easily parallellizable. 

\section{Prediction Quality}

Now that we have two very different algorithms for creating box drawing classifiers, we will compare their performances experimentally.

\subsubsection*{Evaluation Metric}
We chose to use the area under the convex hull of the ROC curve (AUH) \cite{ProvostAUH} as our evaluation metric; it is frequently used for imbalanced classification problems and considers the full ROC curve (Receiver Operator Characteristic) curve to evaluate performance. To compute the AUH, we compute classifiers for various settings of the tradeoff parameter $c$, which controls the relative importance of positive and negative classes. Each setting of $c$ corresponds to a single point on the ROC curve, with a count of true and false positives. We compute the AUH formed by the points on the ROC curve, and normalize as usual by dividing it by the number of positive examples times the number of negative examples. The best possible result is an AUH of 1.

\subsubsection*{Baseline Algorithms}
For comparison, we consider logistic regression, SVM with radial basis kernel, CART, C4.5, Random Forests, AdaBoost (with decision trees), C5.0, and Hellinger Distance Decision Tree (HDDT) \cite{Hellinger}. Most of these algorithms are listed among the top 10 algorithms in data mining \cite{top10}. Among these algorithms, only CART, C4.5, C5.0, and HDDT yield potentially interpretable models. HDDT uses Hellinger distance as the splitting criterion, which is robust and skew-insensitive.

In addition to the baselines above, we implemented the Patient Rule Induction Method (PRIM)  for ``bump hunting" \cite{Friedman}. This method also partitions the input variable space into box shaped regions, but in a different way than our method. PRIM searches iteratively for sub-regions where the target variable has a maxima, and peels them off one at a time, whereas our clustering step finds maxima simultaneously.

The data sets we considered are listed in Table \ref{data}.
Some data sets (castle, corner, diamond, square, flooded, castle3D, corner3D, diamond3D, flooded3D, flooded3D) are simulated data that are able to be visualized (made publicly available at \cite{mypage}). The breast and pima data sets were obtained from the UCI Machine Learning Repository \cite{Bache}. The data set fourclass was obtained from LIBSVM \cite{Chang}. The remaining imbalanced data sets were obtained from the KEEL (Knowledge Extraction based on Evolutionary Learning) imbalanced data repository \cite{KEEL}. The Iris0 data set is an imbalanced version of the standard iris data set, where two of the classes (iris-versicolor and iris-virginica) have been combined to form the majority class. 

\begin{table}[!h]
\small
\centering
\begin{tabular}{|c|p{1.5cm}|p{1.5cm}|p{1.5cm}|}
\hline 
Data &  number of examples & feature size & imbalance ratio \\ \hline
pima & 768 & 8 & 1.8657\\ \hline
castle & 8716 & 2 & 22.2427 \\ \hline
corner &  10000 & 2 & 99\\ \hline
diamond &  10000 & 2 & 24.9067\\ \hline
square & 10000 & 2 & 11.2100\\ \hline
flooded & 10000 & 2 & 31.1543\\ \hline
fourclass & 862 &2 &  1.8078 \\ \hline
castle3D & 545 & 3 & 7.2576\\ \hline
corner3D &  1000 & 3 &28.4118 \\ \hline
diamond3D & 1000 & 3 & 33.4828\\ \hline
square3D &  1000 & 3 & 7\\ \hline
flooded3D & 1000 & 3 & 26.7778\\ \hline
breast & 569 & 30 & 1.6840\\ \hline
abalone19 & 4174 & 9 & 129.4375 \\ \hline
yeast6 &  1484 & 8 & 41.4 \\ \hline
yeast5 & 1484 & 8 & 32.7273\\ \hline
yeast1289 & 947 & 8 & 30.5667\\ \hline
yeast4 &  1484 & 8 & 28.0980\\ \hline
yeast28 &  482 & 8 & 23.1000\\ \hline
yeast1458 & 693 & 8 & 22.1000\\ \hline
abalone918 & 731 & 9 & 16.4048\\ \hline
pageblocks134 & 472 & 10 & 15.8571\\ \hline
ecoli4 & 336 & 7 & 15.8000\\ \hline
yeast17 & 459 & 7 & 14.3\\ \hline
shuttle04 & 1829 & 9 & 13.8699\\ \hline 
glass2 & 214 & 9 & 11.5882 \\ \hline
vehicle3 & 846 & 18 & 2.9906\\ \hline
vehicle1 & 846 & 18 & 2.8986 \\ \hline
vehicle2 &  846 & 18 & 2.8807\\ \hline
haberman &  306 & 3 & 2.7778\\ \hline
yeast1 &  1484 & 8 & 2.4592\\ \hline
glass0 & 214 & 9 & 2.0571  \\ \hline
iris0 & 150 & 4 & 2 \\ \hline
wisconsin & 683 & 9 &  1.8577 \\ \hline
ecoli01 & 220 & 7 & 1.8571 \\ \hline
glass1 & 214 & 9 & 1.8158\\ \hline
breast tissue & 106 & 9 & 3.8182 \\ \hline
\end{tabular}
\caption{Summary of datasets used for experiments}
\label{data}
\end{table}

\subsubsection*{Performance analysis}
Here we compare the performance of Fast Boxes with the baseline algorithms. 
For each algorithm (except C4.5) we set the imbalance weighting parameter to each value $[0.1, 0.2, 0.3,$\\$\ldots,1]$.  The other parameters were set in data-dependent way; for instance, for SVM with RBF kernel, the kernel width was chosen using the sigest function in the R programming language. The data were separated into 10 folds, where each fold was used in turn as the test set. We do not prune the decision trees beyond their built-in pruning as previous research shows that unpruned decision trees are more effective in their predictions on the minority class \cite{ProvostDomingos,Chawla}, and because it would introduce more complexity that would be difficult to control for. Within the training set, for the Fast Boxes algorithm we used 3-fold cross-validation to select the cluster number and expansion parameter.

Table \ref{firstresult1} shows the performances in terms of AUH means and standard deviations. The values that are bolded represent the algorithms whose results are not statistically significantly different from the best algorithm using a matched pairs sign test with significance level $\alpha=0.05$.  
When there was more than one best-performing classifier, the one with the smaller standard deviation was chosen as the best performer for that data set. Fast Boxes was often (but not always) one of the best performers for each dataset. This brings up several questions, such as: \textit{Under what conditions does Fast Boxes perform well? How do its parameters effect the result? Does it tend to produce trivial results? Can Exact Boxes improve upon Fast Boxes' results in cases where it does not perform well? Are the results interpretable?} These are questions we will address in the remainder of this section.

\begin{figure}[!h]
\vspace{0.5cm}
	\centering
		\includegraphics[width=5cm]{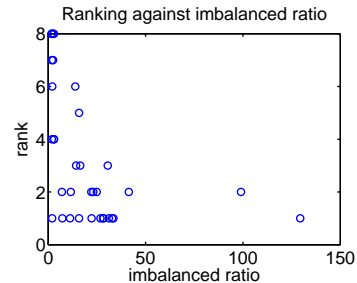}
	\caption{Ranking of Fast Boxes versus imbalance ratio of data}
	\label{fig:ranking}
\end{figure}

We start with a partial answer to the question of when Fast Boxes performs well - it is when the classes are more imbalanced. Figure \ref{fig:ranking} shows a scatter plot of the quality of Fast Boxes' performance versus the imbalance ratio of the dataset. The vertical axis represents our rank in performance among all of the algorithms we tested. The horizontal axis is the number of negatives divided by the number of positives. The performance of Fast Boxes changes from being among the worst performers when the data are not imbalanced (and the cluster assumption is false), to being among the best performers when the data are imbalanced.

Below we provide some intuition about Fast Boxes' clusters and the expansion parameter before answering the questions posed just above.
\subsubsection*{Effect of Fast Boxes' parameter settings}
\begin{figure}[!h]
\vspace{0.5cm}
	\centering
		\includegraphics[width=5cm]{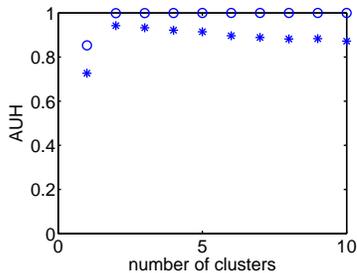}
	\caption{The effect of the number of clusters on AUH for the data set diamond3D. Fast Boxes was run once for each number of clusters. Training AUH is reported as circles, and testing AUH as stars.
	\label{fig:clusterdiamond3d}}
\end{figure}

We expect that if our main modeling assumption holds, which is that the positive examples naturally cluster, there should be a single best number of clusters. If we choose the number of clusters too small, we might underfit, and if we allow too many clusters, we could overfit. Figure \ref{fig:clusterdiamond3d} illustrates the cluster assumption on the diamond3D dataset, where this effect of overfitting and underfitting can be seen.

\begin{figure}[!h]
\vspace{0.5cm}
	\centering
		\includegraphics[width=5cm]{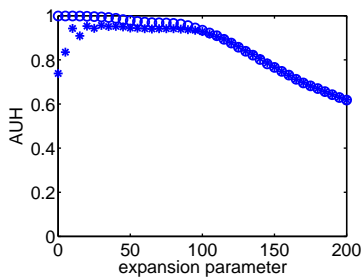}
	\caption{The effect of the expansion parameter on AUH for the diamond3D data set.}
	\label{fig:expanddiamond3d}
\end{figure}

The expansion parameter is also designed to assist with generalization. We would like our boxes to be able to capture more of the positive cluster than is provided by the tightest box around the training examples, particularly since true positives are worth more than true negatives in our objective function. The exponential loss creates a discriminative classifier, but with a push outwards. 
Naturally, as we increase the expansion parameter, the training AUH will drop as more negative training examples are included within the box. On the other hand, the test AUH tends to increase before decreasing, as more positive examples are within the expanded box. This effect is illustrated in Figure \ref{fig:expanddiamond3d}. 

\begin{figure}
\vspace{0.5cm}
	\centering
		\includegraphics[width=6cm]{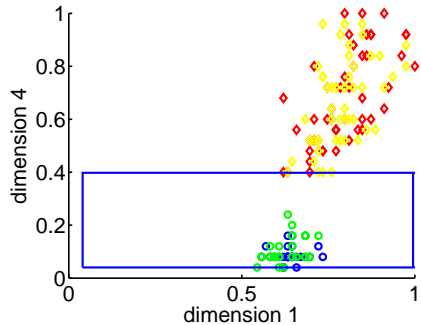}
	\caption{The red and yellow points are negative training points and testing point respectively, the blue and green points are positive training points and testing points respectively. If we had used the tightest decision boundary around the positive training examples, we would have missed part of the positive distribution.}
	\label{fig:minimalexpansion}
\end{figure}

Considering the final expansion stage,
Figure \ref{fig:minimalexpansion} illustrates why this stage is necessary. We visualize the iris0 dataset with dimension 1 and dimension 4, where if we had not expanded out to the nearest negative example, we would have missed a part of the positive distribution within the test set.

\subsubsection*{Production of trivial rules}

When the data are highly imbalanced, we have found that some of the baseline algorithms for producing interpretable models often produce trivial models, that is, models that always predict a single class. This is true even when the weighting factor on the positive class is varied throughout its full range at a reasonably fine granularity. This means that either it is not possible to obtain a meaningful model for the dataset with that method, or it means one would need to work fairly hard in order to find a weighting factor that did not produce a trivial model; that is, the range for which nontrivial models are possible is very small. Figure \ref{trivial rule} considers three interpretable methods we compare with, namely CART, C4.5, and C5.0. It shows the fraction of time these algorithms produce trivial models. For CART, C5.0, and Fast Boxes, the percentage was computed over 100 models computed over 10 splits and 10 options for the imbalance parameter. C4.5 does not have a built in imbalance parameter, so the percentage was computed over 10 splits. 

\begin{table}[!h]
\centering
\small
\begin{tabular}{|c|c|c|c|c|c|}
\hline 
Data &  CART & C4.5 & C5.0 & Fast Boxes \\ \hline
pima & 0.00 & 0.00 & 0.00 & 0.07\\ \hline
castle & 0.00 & 0.00 & 0.00 & 0.10\\ \hline
corner & 0.00 & \textbf{0.60} & \textbf{0.70} & 0.00\\ \hline
diamond & 0.00 & 0.00 & 0.00 & 0.00\\ \hline 
square & 0.00 & 0.00 & 0.00 & 0.00\\ \hline
flooded & 0.00 & \textbf{0.70} & \textbf{0.80} & 0.00\\ \hline
fourclass & 0.00 & 0.00 & 0.00 & 0.03\\ \hline
castle3D & 0.00 & 0.00 & 0.00 & 0.10\\ \hline
corner3D &  0.00 & \textbf{0.50} & \textbf{0.50} & 0.07 \\ \hline
diamond3D & 0.00 & \textbf{1.00} & \textbf{1.00} & 0.06\\ \hline
square3D &  0.00 & \textbf{0.90} & \textbf{0.80} & 0.10\\ \hline
flooded3D & 0.05 & \textbf{1.00} & \textbf{1.00} & 0.09 \\ \hline
breast &  0.00 & 0.00 & 0.00 & 0.37\\ \hline
abalone19 & 0.43 & \textbf{1.00} & \textbf{1.00} & 0.35\\ \hline
yeast6 &  0.00 & 0.00 & 0.00 & 0.02 \\ \hline
yeast5 & 0.00 & 0.00 & 0.00 & 0.36\\ \hline
yeast1289 & 0.16 & \textbf{0.70} & \textbf{0.60} & 0.35\\ \hline
yeast4 &  0.00 & 0.20 & 0.20 & 0.30\\ \hline
yeast28 &  0.39 & \textbf{0.90} & \textbf{0.90} & 0.00\\ \hline
yeast1458 & 0.24 & \textbf{0.70} & \textbf{0.90} & 0.19\\ \hline
abalone918 & 0.00 & 0.10 & 0.10 & 0.40\\ \hline
pageblocks134 & 0.00 & 0.00 & 0.00 & 0.39 \\ \hline 
ecoli4 & 0.00 & 0.00 & 0.00 & 0.32 \\ \hline
yeast17 & 0.03 & 0.20 & 0.30 &  0.21\\ \hline
shuttle04 & 0.00 & 0.00 & 0.00 & 0.00\\ \hline
glass2 & 0.09 & 0.40 & \textbf{0.70} & 0.28 \\ \hline
vehicle3 & 0.00 & 0.00 & 0.00 & 0.01\\ \hline
vehicle1 & 0.00 & 0.00 & 0.00 & 0.02\\ \hline
vehicle2 &  0.00 & 0.00 & 0.00 & 0.27\\ \hline
haberman & 0.00 & 0.40 & \textbf{0.70} &  0.13\\ \hline
yeast1 & 0.00 & 0.00 & 0.00 & 0.08\\ \hline
glass0 & 0.00 & 0.00 & 0.00 & 0.08\\ \hline
iris0 & 0.00 & 0.00 & 0.00 & 0.03\\ \hline
wisconsin & 0.00 & 0.00 & 0.00 & 0.34\\ \hline
ecoli01 & 0.00 & 0.00 & 0.00 & 0.21\\ \hline
glass1 & 0.00 & 0.00 & 0.00 & 0.16 \\ \hline
breast tissue & 0.00 & 0.00 & 0.00 & 0.08 \\ \hline
\end{tabular}
\caption{Fraction of the time we get a trivial model. Bold indicates values over 0.5.}
\label{trivial rule}
\end{table} 

\subsubsection*{Comparison of Fast Boxes and Exact Boxes}
Since we know that Fast Boxes is competitive with other baselines for handling imbalanced data, we would like to know whether Exact Boxes has the potential to yield significant performance gains over Fast Boxes and other methods. We implemented the MIP using GUROBI on a quad core Intel i7 860 2.8 GHz, 8GB cache, processor with 4 cores with hyperthreading and 16GM of RAM.  We first ran the Exact Boxes algorithm for 30 minutes, and if the AUH performance was not competitive and the optimality gap was above $1\%$, we ran it up to 90 minutes for each instance. We did not generally allow the MIP to solve to provable optimality. This has the potential to hinder performance, but as we were performing repeated experiments we needed to be able to solve the method repeatedly.

Table \ref{MIPresult} shows results from Exact Boxes for several of the smaller data sets, along with the results from Fast Boxes for comparison. Bold font in this table summarizes results from the other baseline algorithms as well: if the entry is in bold, it means that the result is not statistically significantly different than the best out of \textit{all} of the algorithms. Thus, for 5 out of 8 datasets we tried, the MIP was among the top performers. Further, the AUH value was substantially improved for some of the data sets. Thus, restricting the algorithm to produce a box drawing classifier does not generally seem to hinder performance.

\begin{table}[!h] 
\small
\centering
\begin{tabular}{|p{1.2cm}|p{1cm}|p{1cm}|p{1cm}|p{1cm}|}
\hline 
Data &  Best Performance  & Fast Boxes & Exact Boxes & Exact Boxes ranking\\ \hline
vehicle2 & 0.9496 (0.015)   & 0.9191 (0.0242) & 0.9496 (0.015) & 1\\ \hline 
haberman  & 0.6699 (0.0276)& 0.5290 (0.0265)   &  \textbf{0.6632} (0.0303) & 2\\ \hline
yeast1 & 0.7641 (0.0133)  & 0.5903 (0.0286)  & 0.7392 (0.0172) & 2\\ \hline 
glass0  & 0.8312 (0.0345) & 0.7937 (0.0212)  & 0.7977 (0.0421) & 2 \\ \hline 
iris0 &  1 \newline (0)  &  \textbf{1} \newline (0)  & \textbf{1} \newline (0) & 1\\ \hline
wisconsin & 0.9741 (0.0075)  &  0.8054 (0.1393)  & \textbf{0.9726} (0.0079) & 2\\ \hline
ecoli01 & 0.9840 (0.0105)  &  0.9433 (0.0300)& \textbf{0.9839} (0.0109) & 2\\ \hline  
glass1 & 0.7922 (0.0377)  & 0.6654 (0.0356)&\textbf{0.7922} (0.0337) & 1\\ \hline 
\end{tabular}
\caption{Comparison of test data AUH of interpretable methods with Exact Boxes. Bold font includes results from non-interpretable methods.}
\label{MIPresult}
\end{table} 


Note that it is time-consuming to perform cross-validation on the MIP, so the cluster number that we found using cross-validation for Fast Boxes was used for Exact Boxes.

%
%

\subsubsection*{Interpretability demonstration} 
We provide a classifier we learned from the glass2 data set that predicts whether a particular glass is a building window that is non-float processed. The other types of glasses are building windows that are float processed, vehicle windows, containers, tableware, and headlamps. The attributes include the refraction index as well as various indices for metals. These metals include Sodium, Magnesium, Aluminum, Silicon, Potassium, Calcium, Barium, and Iron. 

One of the predictive models from Fast Boxes is as follows.
To be a particular glass of a building window that is non-float processed:\\
1) The refractive index should be above 1.5161.\\
2) Magnesium index must be above 3.3301.\\
3) Aluminum should be below 1.7897.\\
4) Silicon should be below 73.0199.\\
5) Potassium should be below 0.6199.\\
6) Calcium should be between 8.3101 and 2.3741.\\
7) Barium should be below 2.6646.\\
8) Sodium and iron are not important factors. \\

We believe that this simple form of model would appeal to practitioners because of the natural threshold structure of the box drawing classifiers.

\section{Theoretical guarantee on performance}

Statistical learning theory will allow us to provide a probabilistic guarantee on the performance of our algorithms. We will construct a uniform generalization bound, which holds over all box drawing classifiers with $K$ boxes anchored at $M_j$ different fixed values for each dimension, where $K$ is fixed. We might choose $M_j$ as the count of numbers with at most a certain number of decimal places (say 2 decimal places) in between the largest and smallest possible values for a particular feature. (Often in practice only 2 decimal places are used.) The main step in our proof is to count the number of possible box drawing classifiers. The set of all box drawing classifiers with up to $K$ boxes, with $l_j$ and $u_j$ attaining the $M_j$ values, will be called $F$.

Define the empirical risk to be the objective of Exact Boxes with no regularization, 
$$R^{emp}(f)=\sum_{i:y_i=1} \one_{[f(\mathbf{x}_i)=1]} + C_I \sum_{i:y_i=-1} \one_{[f(\mathbf{x}_i)=-1]} ,$$
and let the true risk $R^{true}(f)$ be the expectation of this taken over the distribution that the data are drawn iid from. 
 
\begin{proposition}
For all $\delta>0$ with probability at least $1-\delta, \forall f \in F$, $$R^{true}(f) \leq R^{emp}(f)+\sqrt{\frac{K \sum_{j=1}^n \log ( \frac{M_j(M_j-1)}{2})-log K! +\log \frac{1}{\delta}}{2m}}.$$
\end{proposition}

To outline the proof, there are $ \prod_{j=1}^n\left( \begin{array}{c} M_j \\ 2\end{array} \right)$ ways to construct a single box, since for each dimension, we select 2 values, namely the lower boundary $l_j$ and upper boundary  $u_j$. To construct multiple boxes, there are at most $ \prod_{j=1}^n\left( \begin{array}{c} M_j \\ 2\end{array} \right)^K$ ways if the order of construction of the boxes matter. Since the order does not matter, we need to divide the term by $K!$. Note that this is an upper bound which is not tight since some boxes can be a proper subset or equal to another box. 
Although we are considering the set of all box drawing classifiers up to $K$ boxes, it suffices to consider box drawing classifiers with exactly $K$ boxes. This can be seen by supposing we constructed a classifier with $l<K$ boxes, and noting the same classifier can be constructed using $K$ boxes by duplicating some boxes. We apply Hoeffding's inequality and the union bound to complete the proof.

\section{Making the MIP more practical}

From the experimental outcome, it is clear that Exact Boxes is indeed a competitive solution. The main challenge lies in its computational complexity. There are several ways one might make the MIP more practical: first, one could limit computation to focus only a neighborhood of the positive data, and use the solution to this problem to warm start the MIP on the full problem. In that case we would consider only negative points that are close to the positive points in at least one dimension, which can be identified in a single pass through the negative examples. 
Alternatively, one can perform clustering first as in the Fast Boxes approach, and solve the MIP on each cluster. For each cluster, we would scan through each feature of the data in a single pass and keep only the data that are close to the mean of the cluster center to use in the MIP. 

\section{Discussion and Conclusion}

We have presented two new approaches to designing interpretable predictive models for imbalanced data settings. Exact Boxes is formulated as a mixed integer program, and acts as a gold standard interpretable modeling technique to compare with. It can be used for small to moderately sized problems. Fast Boxes uses a characterize-then-discriminate approach, and tends to work well when the minority class is naturally clustered (for instance when the clusters represent different failure modes of mechanical equipment). We illuminated the benefits and limitations of our approaches, and hope that these types of models will be able to provide alternative explanations and insights into imbalanced problems. 
In comparing Fast Boxes with gold standard interpretable techniques like Exact Boxes, and with many other methods, we can now judge the power of the class of interpretable models: it is interesting that such simple approaches can achieve comparable performance with even the best state-of-the-art techniques. \\

\noindent \textbf{Acknowledgements} Funding for this work provided by Siemens.

\bibliographystyle{tech}
\bibliography{fp0491-gohbib}

\begin{table*}[!h] 
\tiny
\centering
\tabcolsep=0.11cm
\begin{tabular}{|p{1.3cm}|p{0.95cm}|p{0.95cm}|p{0.95cm}|p{0.95cm}|p{0.95cm}|p{0.95cm}|p{0.95cm}|p{0.95cm}|p{1cm}|}
\hline 
Data &  Logistic       & SVM             & CART            &C4.5             & Ada- \newline Boost         & RF & C5.0 &  HDDT & Fast Boxes\\ \hline
pima & \textbf{0.8587} \newline (0.0112) & \textbf{0.8468} (0.0126) & 0.7738 (0.0123) & 0.6579 (0.0347) & 0.6810 (0.0218) & 0.6942 (0.0126) & 0.6574 (0.0353) &  0.6642 (0.0274) & 0.7298 (0.0241)\\ \hline 
castle & 0.5 \newline (0) & \textbf{1} \newline (0) & 0.9941 (0.0068) & 0.9947 (0.0060) & \textbf{0.9949} (0.0046) & \textbf{0.9922} (0.0079) & 0.9941 (0.0060) & \textbf{0.9949} (0.0062)& \textbf{1} \newline (0)\\ \hline
corner & \textbf{0.9871} (0.0129) & \textbf{0.9948} (0.0005) & 0.9488 (0.2717) & 0.5997 (0.1482) & 0.6984 (0.0449) & 0.6828 (0.0265) & 0.5612 (0.1110)  & 0.6865 (0.0365) & 0.9891 (0.0001)\\ \hline
diamond & 0.5 \newline (0) & \textbf{0.9980} (0.0004) & 0.9585 (0.0129) & 0.9328 (0.0181) & 0.9460 (0.0117) & 0.9433 (0.0121) & 0.9311 (0.0208) & 0.9364 (0.0180)& 0.9744 (0.0062)\\ \hline
square & 0.5404 (0.0718) & 0.9944 (0.0001) & 0.9949 (0.0051) & 0.9949 (0.0043) & 0.9939 (0.0033) & 0.9947 (0.0033) & 0.9949 (0.0043)  & 0.9949 (0.0027) & \textbf{0.9984} (0.0015)\\ \hline
flooded & 0 \newline (0) & \textbf{0.9831} (0.0010) & 0.9466 (0.0157) & 0.5488 (0.1074) & 0.7017 (0.0231) & 0.7036 (0.0252) & 0.5482 (0.1077)  & 0.6992 (0.0208) & 09638 (0.0091)\\ \hline
fourclass &  0.8122 (0.0195) & \textbf{0.9957} (0.0176) & 0.9688 (0.0176) & 0.9916 (0.0296) & 0.9670 (0.0265) & \textbf{0.9920} \newline (0.0053) & 0.9670 (0.0130)  & 0.9698 (0.0116) & 0.9546 (0.0174)\\ \hline
castle3D & 0.5449 (0.0324) & \textbf{1} \newline (0) & 0.9532 (0.0347) & 0.9530 (0.0374) & 0.9272 (0.0499) & \textbf{0.9455} (0.0563) & 0.9439 (0.0615)  & 0.9530 (0.0374) & \textbf{1} \newline (0)\\ \hline
corner3D & 0.8448 (0.0316) & \textbf{0.9225} (0.0463) & 0.8481 (0.0504) & 0.5596 (0.0729) & 0.6245 (0.03927) & 0.5657 (0.0309) & 0.5622 (0.0778)  & 0.6413 (0.0457) & \textbf{0.9736} (0.0091) \\ \hline
diamond3D & 0.5449 (0.0324) & \textbf{0.7962} (0.0917) & 0.7372 (0.0347) & 0.5 (0.0374) & 0.5492 (0.0499) & 0.5957 (0.0309) & 0.5622 (0.0778)  & 0.6883 (0.0542) &  \textbf{0.9516} (0.0119)\\ \hline
square3D & 0.5 \newline (0) & \textbf{0.9626} (0.0156) & \textbf{0.9106} (0.0306) & 0.5387 (0.1224) & 0.8703 (0.01451) & 0.8790 (0.0234) & 0.5811 (0.1712)  & 0.9034 (0.0322) & \textbf{0.9578} (0.0090)\\ \hline
flooded3D & 0.5 \newline (0) & 0.7912 (0.0781) & 0.7724 (0.0902) & 0.5 \newline (0) & 0.5471 (0.0329) & 0.5489 (0.0440) & 0.5 \newline (0)  & 0.6422 (0.0749) & \textbf{0.9233} (0.0307)\\ \hline
breast &  0.9297 (0.0230) & \textbf{0.9801} (0.0079) & 0.9516 (0.0173) & 0.9251 (0.0138) & 0.9457 (0.0329) & 0.9609 (0.0102) & 0.9281 (0.0135) & 0.9231 (0.0180)& 0.8888 (0.0313) \\ \hline
abalone19 & 0.5188 (0.0182) & 0.5 \newline (0) & 0.5382 (0.0261) & 0.5 \newline (0) & 0.5 \newline (0) & 0.5 \newline (0) & 0.5 \newline (0)  & 0.5116 (0.0164) & \textbf{0.6882} (0.0583)\\ \hline
yeast6 & \textbf{0.8503} (0.0341) & \textbf{0.8649} (0.0246) & 0.7995 (0.0624) & 0.7129 (0.0829) & 0.7126 (0.0536) & 0.7277 (0.0581) & 0.7129 (0.0853)  & 0.7064 (0.0772) & \textbf{0.8609} (0.0585)\\ \hline
yeast5 & \textbf{0.9499} (0.0479) & 0.9229 (0.0339) & \textbf{0.9197} (0.0575) & 0.8280 (0.1159) & 0.8305 (0.0859) & 0.8061 (0.0616) & 0.8241 (0.1157)  & 0.7931 (0.1126)& \textbf{0.9767} (0.0092)\\ \hline
yeast1289 & \textbf{0.6319} (0.0433) & 0.5618 (0.0332) & \textbf{0.7076} (0.0665) & 0.5088 (0.0322) & 0.5152 (0.0288) & 0.5067 (0.0141) & 0.5156 (0.0342)  & 0.5531 (0.0436) & 0.5932 (0.0557)\\ \hline
yeast4 & 0.8001 (0.0309) & 0.7836 (0.0480) & 0.7595 (0.0410) & 0.6115 (0.0902) & 0.6131 (0.0326) & 0.5922 (0.0326) & 0.6210 (0.07899) & 0.6289 (0.0471) & \textbf{0.8794} (0.0274) \\ \hline
yeast28 & \textbf{0.7907} (0.0525) & 0.6596 (0.0565) & 0.6402 (0.0893) & 0.5100 (0.0316) & 0.5 \newline (0) & 0.6489 (0.0472) & 0.5248 (0.0784)  & 0.6126 (0.0606) & \textbf{0.7366} (0.0467)\\ \hline
yeast1458 & \textbf{0.6164} (0.0510) & 0.5420 (0.0322) & \textbf{0.6032} (0.0281) & 0.5 \newline (0) & 0.5023 (0.0088) & 0.5095 (0.0154) & 0.5 \newline (0)  & 0.5340 (0.0467) & \textbf{0.6090} (0.0431)\\ \hline
abalone918 & \textbf{0.8849} (0.0270) & 0.6780 (0.0391) & 0.7427 (0.0517) & 0.5904 (0.0581) & 0.6117 (0.0456) & 0.5580 (0.03213) & 0.5725 (0.0470) & 0.6310 (0.0418) & 0.7171 (0.0603)\\ \hline
pageblocks \newline 134 & 0.9461 (0.0444) & 0.7874 (0.1184) & \textbf{0.9945} (0.0109) & \textbf{0.9908} (0.0219) & \textbf{0.9908} (0.0449) & \textbf{0.9500} (0.0345) & \textbf{0.9908} (0.0219)  & \textbf{0.9551} (0.0487) & 0.9500 (0.0359)\\ \hline
ecoli4 & \textbf{0.8926} (0.0615) & \textbf{0.9176} (0.0424) & \textbf{0.8809} (0.0593) & 0.7759 (0.07756) & \textbf{0.7965} (0.0775) & \textbf{0.8494} (0.0775) & 0.8471 (0.0532)  & \textbf{0.8430} (0.0743) & \textbf{0.9202} (0.0622)\\ \hline
yeast17 & \textbf{0.7534} (0.0611) & \textbf{0.6905} (0.0386) & \textbf{0.7481} (0.0713) & 0.5841 (0.0698) & 0.5382 (0.0225) & 0.5529 (0.0359) & 0.5721 (0.0699)  & 0.6070 (0.0509) & \textbf{0.7033} (0.0547)\\ \hline
shuttle04 & \textbf{0.9965} (0.0045) & \textbf{0.9828} (0.0105) & \textbf{1} \newline (0) & \textbf{0.9994} (0.0008) & \textbf{1} \newline (0) & \textbf{1} \newline (0) & \textbf{1} \newline (0) & \textbf{0.9994} (0.0008) & \textbf{0.9967} (0.0042)\\ \hline
glass2 &  \textbf{0.7609} (0.0726) & 0.6128 (0.0941) & \textbf{0.7112} (0.1090) & 0.5541 (0.0640) & 0.5324 (0.0417) & 0.5479 (0.0597) & 0.5200 (0.0415) & 0.5892 (0.0573) & \textbf{0.7334} (0.0904)\\ \hline
vehicle3 & \textbf{0.8397} (0.0079) & \textbf{0.8524} (0.0169) & 0.7733 (0.0255) & 0.6515 (0.0401) & 0.6591 (0.0212) & 0.6484 (0.0232) & 0.6621 (0.02117)  & 0.6823 (0.0300) & 0.7003 (0.0267)\\ \hline
vehicle1 & \textbf{0.8587} (0.0112) & 0.8468 (0.0126) & 0.7738 (0.0123) & 0.6579 (0.0347) & 0.6810 (0.0218) & 0.6942 (0.0126) & 0.6574 (0.0353)  & 0.6719 (0.0265) & 0.7298 (0.0241)\\ \hline
vehicle2 & 0.9632  (0.0134) & \textbf{0.9837} (0.0072) & 0.9437 (0.01880) & 0.9351 (0.0133) & 0.9677 (0.0097) & \textbf{0.9775} (0.0106) & 0.9365 (0.0129)  & 0.9248 (0.0243) & 0.9191 (0.0242)\\ \hline
%
%
haberman & \textbf{0.6589} (0.1713) & \textbf{0.6898} (0.0427) & \textbf{0.6699} (0.0276) & \textbf{0.5733} (0.0748) & 0.6004 (0.0323) & 0.6130 (0.0318) & \textbf{0.5420} (0.6780)  & 0.5604 (0.0231) & 0.5290 (0.0265)\\ \hline 
yeast1 & 0.7836 (0.0184) & \textbf{0.7991} (0.0150) & 0.7641 (0.0133) & 0.6672 (0.0372) & 0.6859 (0.0219) & 0.6130 (0.0318) & 0.5420 (0.0678)  & 0.6369 (0.0128)& 0.5903 (0.0286)  \\ \hline 
glass0 & 0.7951 (0.0437) & \textbf{0.8636} (0.0336) & 0.8312 (0.0345) & 0.7687 (0.0619) & 0.7998 (0.0381) & \textbf{0.8572} (0.0281) & 0.7690 (0.0595)  & 0.7569 (0.0424) & 0.7937 (0.0212)\\ \hline 
iris0 &  \textbf{1} \newline (0) & \textbf{0.998} (0.0063) & \textbf{1} \newline (0) & \textbf{0.978} (0.0175) & \textbf{1} \newline (0) &\textbf{1} \newline (0) & \textbf{0.972} (0.0169) & \textbf{0.9880} (0.0193) & \textbf{1} \newline (0) \\ \hline 
wisconsin & \textbf{0.9746} (0.0093) & \textbf{0.9735} (0.0073) & \textbf{0.9741} (0.0075) & 0.9455 (0.0124) & 0.9611 (0.0122) & 0.9672 (0.0072) & 0.9416 (0.0121)  & 0.9249 (0.0203) & 0.8054 (0.1393)\\ \hline 
ecoli01 & \textbf{0.9728} (0.0140) & \textbf{0.9850} (0.0091) & \textbf{0.9840} (0.0105) & \textbf{0.9806} (0.0107) & \textbf{0.9828} (0.0063) & \textbf{0.9855} (0.0097) & \textbf{0.9806} (0.0107)  & 0.9806 (0.0107) & 0.9433 (0.0300)\\ \hline 
glass1 & 0.7247 (0.0363) & \textbf{0.8057} (0.0340) & 0.7598 (0.0490) & 0.7050 (0.0358) & 0.6997 (0.0478) & 0.7833 (0.0274) & 0.6822 (0.0320) &0.7189 (0.0586) & 0.6654 (0.0356) \\ \hline 
breast tissue & 0.9411 (0.0394) & 0.9908 (0.0064) & 0.9417 (0.0747) & 0.9450 (0.0602) & 0.9632 (0.0297) & 0.9531 (0.0505) & 0.9630 (0.0403) & 0.9314 (0.0550) &\textbf{0.9953} (0.0042) \\ \hline
\end{tabular}
\caption{Comparison of test data AUH of Fast Boxes with other algorithms}
\label{firstresult1}
\end{table*}

\end{document}